# AI Application in Anti-Money Laundering for Sustainable and Transparent Financial Systems

Chuanhao Nie*

College of Computing, Georgia Institute of Technology, Atlanta, United States, cnie30@gatech.edu

Yunbo Liu

Department of Electrical and Computer Engineering, Duke University, Durham, United States, yunbo.liu954@duke.edu

Chao Wang

Department of Computer Science, Rice University, Houston, United States, zylj2020@outlook.com

**Abstract**

Money laundering and financial fraud remain major threats to global financial stability, costing trillions annually and challenging regulatory oversight. Recent research demonstrates that financial compliance—when supported by RegTech, artificial intelligence, and digital automation—can reduce KYC/AML processing time, strengthen customer trust, and contribute to long-term sustainable development outcomes in the digital economy. These findings highlight the increasing strategic importance of technology-enabled compliance and motivate the need for AI-driven solutions that improve efficiency while supporting transparency and responsible governance. This paper reviews how artificial intelligence (AI) applications can modernize Anti-Money Laundering (AML) workflows by improving detection accuracy, lowering false-positive rates, and reducing the operational burden of manual investigations, thereby supporting more sustainable development. It further highlights future research directions including federated learning for privacy-preserving collaboration, fairness-aware and interpretable AI, reinforcement learning for adaptive defenses, and human-in-the-loop visualization systems to ensure that next-generation AML architectures remain transparent, accountable, and robust. In the final part, the paper proposes an AI-driven KYC application that integrates graph-based retrieval-augmented generation (RAG Graph) with generative models to enhance efficiency, transparency, and decision support in KYC processes related to money-laundering detection. To the best of our knowledge, no prior research has combined graph-based RAG architectures with generative AI specifically for KYC Customer Due Diligence (CDD)/Enhanced Due Diligence (EDD) in AML. Experimental results show that the RAG-Graph architecture delivers high faithfulness and strong answer relevancy across diverse evaluation settings, thereby enhancing the efficiency and transparency of KYC CDD/EDD workflows and contributing to more sustainable, resource-optimized compliance practices.

## 1    INTRODUCTION

Money laundering and financial fraud continue to undermine global financial stability and public trust. The United Nations Office on Drugs and Crime estimates that 2–5 percent of global GDP—about US $800 billion to $2 trillion—is laundered annually [1], while recent UN data suggest around US $1.6 trillion (≈ 2.7 percent of GDP) is involved [2]. Fraud in digital payments is also rising: the Nilson Report recorded US $33.83 billion in global card-fraud losses in 2023, with the U.S. responsible for 42 percent [3]. Under tightening regulatory pressure, financial institutions must strengthen transaction monitoring, KYC, and SAR processes. The Financial Action Task Force warns that cyber-enabled fraud increasingly drives illicit financial flows, calling for stronger cross-border cooperation and modernized compliance systems [4]

Traditional compliance systems rely heavily on rule-based monitoring, threshold scenarios, and manual investigations. While these approaches provide a baseline level of compliance, they are plagued by excessive false positives, high operational costs, and limited adaptability to evolving criminal typologies such as trade-based money laundering and emerging channels involving new financial instruments [5,6]. This rigidity has created a significant gap between regulatory expectations and the actual effectiveness of AML and fraud detection programs. Moreover, inefficient compliance workflows can increase resource consumption, generate operational waste, and reduce the transparency and trust needed for sustainable financial ecosystems. Recent research shows that technology-enhanced

compliance—particularly when supported by AI and RegTech—strengthens organizational resilience, improves customer trust, and contributes to broader sustainable development outcomes in the digital economy[33].

Artificial intelligence (AI) and machine learning (ML) are increasingly viewed as transformative solutions to these challenges. Advances in anomaly detection [7], graph-based learning [8], natural language processing [9], and deep learning architectures [10] have opened new possibilities for uncovering hidden laundering patterns, reducing false positives, and automating compliance workflows. At the same time, emerging technologies such as federated learning and privacy-preserving computation promise to reconcile the tension between cross-institutional intelligence sharing and strict data-protection requirements [11]. In parallel, advances in LLM tooling are making SAR and KYC automation more feasible in practice. Zeng et al.'s *FineEdit* (2025) further demonstrates the potential of precise, regulator-controllable model editing for compliant SAR/KYC template updates and error correction [25].

This paper aims to review the current progress of AI applications in anti–money laundering and fraud detection, focusing on their demonstrated impact on accuracy and efficiency across transaction monitoring, fraud prevention, SAR reporting, and KYC. It also examines the constraints that limit adoption, including data scarcity, explainability, fairness, and regulatory trust. Finally, the paper outlines future research directions that can guide the development of next-generation AML systems, emphasizing the need for approaches that are not only technically advanced but also transparent, privacy-preserving, and aligned with compliance obligations. As financial institutions increasingly integrate sustainability principles into risk management, AI-enabled compliance systems offer an opportunity to reduce operational waste, strengthen transparency, and improve customer trust—factors shown to contribute directly to sustainable development in the digital economy [33]. Embedding sustainability considerations into AI-driven AML design ensures that these systems not only enhance detection performance but also support long-term institutional resilience and responsible governance.

## 2  Traditional AML/Fraud Detection Approaches

Anti–money laundering (AML) programs have historically been dominated by rule-based systems. Transaction monitoring frameworks relied on threshold-based alerts—for example, flagging cash deposits above $10,000 or wire transfers that exceeded predefined limits. These scenarios were often developed in consultation with regulators and compliance officers, but once codified, they remained static and failed to evolve with changing laundering typologies. As studies have noted, criminals quickly learned to evade such controls by "structuring" transactions just below thresholds or dispersing funds across multiple accounts, leaving banks with an overwhelming number of false positives and low detection recall [6].

The data architecture underpinning these systems also contributed to inefficiency. Traditional implementations relied on relational databases (RDBMS), where transaction monitoring required repeated joins across multiple tables such as accounts, customers, and transaction histories. While RDBMS offered stable storage and query capabilities, they struggled with the networked nature of laundering, where money moves across layers of accounts in complex cycles. Literature comparing relational and graph databases shows that while relational queries are efficient for single-table retrieval, they become computationally expensive and opaque when used for multi-hop transaction tracing [12].

Customer due diligence (CDD) and Know Your Customer (KYC) processes were similarly limited by their reliance on static customer profiles. Information collected during onboarding—such as identity documents, nationality, or declared occupation—was rarely updated dynamically. As a result, risk ratings could remain unchanged for years, failing to capture behavioral shifts such as sudden changes in transaction volume or geography. This rigidity was highlighted in Fighting Money Laundering with Technology: A Case Study of Bank X in the UK, where legacy systems produced a backlog of outdated profiles and inconsistent case escalation [5].

Fraud detection in earlier generations of compliance programs relied largely on manual auditing and tabular reviews. Investigators would scan reports for unusual transactions or patterns, often supported by simple statistical rules such as frequency counts or ratio tests. While adequate at small scale, these methods proved ineffective as digital payments, credit card transactions, and cross-border flows surged in volume. Visualization-based approaches began to appear later [13], but early systems lacked the scalability to handle the complexity of real-world fraud networks.

Taken together, these traditional approaches suffered from systemic weaknesses: false positives that overwhelmed investigators, poor scalability in the face of massive transaction datasets, and a lack of adaptability to new laundering or fraud typologies. These limitations created the conditions that made AI and machine learning attractive, not as incremental add-ons, but as transformative solutions capable of reshaping AML and fraud detection.

## 3    CURRENT PROGRESS OF AI APPLICATION

To illustrate the integration of artificial intelligence across the AML lifecycle, Figure 1 presents an overview of a modern AI-enabled compliance pipeline. The process begins with data ingestion from heterogeneous sources such as transaction records, KYC databases, and external watchlists. These inputs undergo feature engineering and data normalization before entering various machine-learning modules, including supervised classifiers, graph-based anomaly detectors, and natural language processing components. The outputs are refined through explainability layers and retrieval-augmented generation (RAG) to support investigator review and automated Suspicious Activity Report (SAR) drafting. This architecture highlights how AI components interoperate to enhance both accuracy and operational efficiency across transaction monitoring, fraud detection, and regulatory reporting.

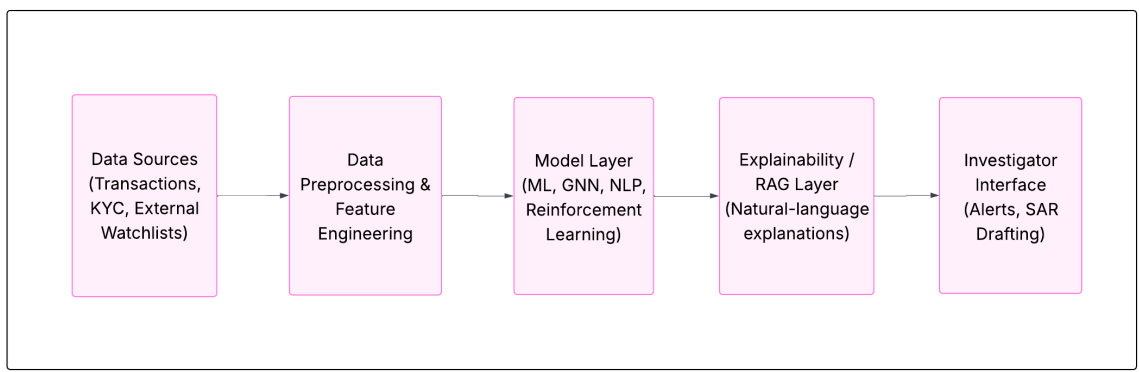

Figure 1 Overview of a Modern AI-enabled Compliance Pipeline

### 3.1    Transaction Monitoring

Transaction monitoring is the central pillar of AML compliance, and also the area where artificial intelligence has made the most visible progress. Traditional systems based on static thresholds and rules generated excessive false positives while failing to capture subtle or cross-entity laundering behavior. To address these shortcomings, researchers have applied supervised and unsupervised machine learning to large transaction datasets. Early studies demonstrated how methods such as decision trees, support vector machines, and random forests could outperform handcrafted scenarios in distinguishing between normal and suspicious behavior. For example, Machine Learning-Based Pattern Recognition for Anti-Money Laundering in Banking Systems [10] combined XGBoost, LSTMs, and Isolation Forest anomaly detection in a hybrid ensemble, achieving an F1 score of 0.91 with false positives below 3%, compared to >95% in rule-based systems. These results illustrate how AI can significantly improve both accuracy and operational efficiency in monitoring.

A major challenge in supervised learning is class imbalance, since confirmed suspicious cases make up less than 1% of all transactions. To overcome this, semi-supervised and imbalanced-learning techniques have been proposed. Leveraging Machine Learning in the Global Fight Against Money Laundering and Terrorism Financing: An Affordances Perspective [14] argued that ML's primary affordance in AML is the ability to cluster unusual activity and flag anomalies, even when labeled data is scarce. Similarly, In Threshold Fine-Tuning of Money Laundering Scenarios Through Multi-Dimensional Optimization Technique, the authors combined K-means clustering and optimization to fine-tune AML scenario thresholds. Their two-stage approach reduced alert volume by 5.4% and unique customer alerts by 1.6% without sacrificing suspicious activity detection [6]. These approaches illustrate a shift from purely supervised learning toward adaptive, semi-supervised pipelines that integrate anomaly detection into monitoring workflows.

The most transformative advances have come from graph-based approaches, which treat financial transactions as dynamic networks rather than isolated records. Scalable Semi-Supervised Graph Learning Techniques for Anti-Money Laundering [8] framed monitoring as a graph learning problem, using models such as EvolveGCN and GraphSAGE to detect laundering motifs like gather–scatter and circular layering. These methods outperformed pipeline ML by capturing temporal dynamics across accounts and institutions. Similarly, Regulatory Graphs and GenAI for Real-Time

Transaction Monitoring and Compliance Explanation in Banking [15] combined GNN-based classification with retrieval-augmented generation (RAG) to automatically align alerts with regulatory requirements and generate natural-language justifications. Such graph-based frameworks not only boost accuracy—achieving F1 scores above 98% in benchmark datasets—but also improve explainability and regulatory trust, addressing a key adoption barrier for AI in compliance.

Taken together, these advances show a clear progression: from threshold-based alerts to supervised ML, from anomaly clustering to semi-supervised learning, and from individual transaction models to graph-based systems that capture systemic laundering behavior. Transaction monitoring is thus evolving from a reactive compliance task into a real-time, adaptive, and explainable AI-driven capability at the core of financial integrity.

### 3.2  Fraud Detection

Fraud detection shares many challenges with AML monitoring, particularly the need to detect rare, adaptive, and high-impact events in massive transaction streams. Traditional approaches relied on manual auditing and static statistical tests, which were quickly overwhelmed by the scale of digital payments and credit card transactions. These methods produced limited coverage of fraudulent schemes and struggled to adapt as criminals evolved tactics, leaving financial institutions with significant losses and operational inefficiencies [13].

The introduction of AI has enabled pattern recognition techniques that are well-suited for credit card and e-commerce fraud. Supervised learning models such as logistic regression, decision trees, and neural networks have been applied to millions of labeled transactions, identifying subtle temporal and behavioral features (e.g., frequency of small purchases, sudden geographic shifts) that distinguish fraud from legitimate use. Ensemble approaches such as random forests and boosting methods further improve robustness by capturing complex feature interactions.

In addition to traditional ML classifiers, visualization-based anomaly detection has become an important tool for fraud analysts. Visualization techniques—scatter plots, bar charts, parenclitic networks, and graph-based outlier detection—provide interpretable insights into suspicious clusters and relationships that may be opaque in black-box models. Lokanan [13] highlights how visual analytics enhances knowledge discovery by exposing hidden connections, such as shared devices or merchants across fraudulent accounts, that automated classifiers may overlook. By enabling investigators to "see" anomalies, visualization bridges the gap between statistical detection and human interpretability.

More recently, researchers have begun exploring reinforcement learning (RL) for adaptive fraud defense. Unlike supervised models that learn from static training sets, RL agents dynamically update detection policies in response to adversarial behavior. This allows monitoring systems to anticipate evolving fraud strategies, such as adaptive credit card testing or account takeover patterns, rather than reacting only after losses have occurred. In experimental settings, RL-driven fraud detection has demonstrated the ability to minimize false negatives while balancing operational costs, though real-world deployment remains limited due to the complexity of training environments and the need for regulator-acceptable explainability [16].

Together, these developments illustrate how fraud detection has evolved from manual, tabular reviews to AI-driven pattern recognition, visual analytics, and adaptive reinforcement learning. The trajectory underscores a dual imperative: maximizing predictive accuracy while ensuring results remain interpretable and aligned with compliance obligations.

### 3.3  Suspicious Activity Reporting (SAR)

Suspicious Activity Reporting (SAR) is a cornerstone of AML compliance, serving as the formal mechanism by which financial institutions escalate suspected money laundering or fraud to regulators. Traditionally, SARs have been drafted manually by investigators, who compile transaction details, customer information, and narrative justifications into regulator-prescribed formats. While essential for compliance, this process is time-consuming, subjective, and inconsistent, often leading to delays in filing, variable report quality, and missed opportunities to disrupt illicit activity [5].

AI and machine learning have introduced innovations that automate and enhance SAR workflows. Early applications involved natural language processing (NLP) to analyze investigator notes, transaction narratives, and historical SAR filings. These systems helped standardize language, identify common typologies, and reduce duplication across reports [17]. More advanced approaches use NLP to extract key entities (accounts, counterparties, geographies) and relationships from unstructured text, enabling richer SARs that capture the full context of suspicious behavior.

A growing area of research focuses on generative AI and retrieval-augmented generation (RAG) for SAR drafting and regulatory explanation. Regulatory Graphs and GenAI for Real-Time Transaction Monitoring and Compliance

Explanation in Banking [15] proposed a pipeline where flagged transactions are not only detected by graph neural networks but also automatically linked to relevant AML regulations, with natural-language justifications generated for compliance officers. This significantly reduces investigator workload and ensures that SAR narratives are consistent with regulatory expectations, while still requiring human review before submission.

Equally important is the application of explainable AI (XAI) to SAR processes. Models that detect suspicious activity must provide interpretable rationales for why a transaction or account is flagged, both to support investigators and to satisfy regulatory requirements. Techniques such as SHAP values, attention heatmaps, and graph motif explanations are being integrated into SAR drafting systems, offering investigators clear evidence trails that can be directly embedded into reports [18].

These developments indicate a shift from SAR as a manual reporting obligation to SAR as a semi-automated, intelligence-driven process. By leveraging NLP, generative AI, and explainability, institutions can produce higher-quality reports, accelerate submission timelines, and align outputs more closely with regulatory frameworks—while reducing the investigative burden on compliance teams.

### 3.4     Know Your Customer (KYC) and Risk Profiling

Know Your Customer (KYC) and risk profiling represent the foundation of AML frameworks, establishing how clients are identified, verified, and monitored over time. Traditionally, these processes relied on static customer profiles built during onboarding, based on attributes such as nationality, occupation, and declared source of funds. Once established, risk ratings often remained unchanged unless manually updated, leaving institutions blind to shifts in customer behavior that could signal laundering or fraud. This rigidity created outdated risk assessments and contributed to inefficiencies in compliance operations [5].

Recent advances in AI have enabled adaptive risk scoring systems, where profiles evolve continuously as new transaction and relationship data are ingested. By linking fragmented data sources, entity linking methods consolidate accounts, counterparties, and products into unified customer views. Biometric technologies—such as facial recognition and fingerprint verification—have also enhanced KYC processes, particularly during onboarding, reducing identity fraud and strengthening continuous monitoring [17]

Beyond static attributes, behavioral profiling has emerged as a promising approach. Clustering methods and autoencoders can group customers into dynamic peer groups and detect outliers whose activity diverges significantly from the norm. For example, Client Profiling for an Anti-Money Laundering System [19] demonstrated how clustering could refine risk segmentation, prioritizing higher-risk clients for monitoring and investigation. Autoencoders compress behavioral features into latent representations, flagging deviations that may not be visible in raw transactional data but are strongly correlated with suspicious activity.

Another emerging focus in KYC model development is fairness and bias auditing. AI-driven profiling systems may unintentionally reproduce biases embedded in training data—such as over-penalizing customers from certain geographies, demographics, or occupations—raising both ethical and regulatory challenges. Recent research, including AI-Driven Fraud Detection and Biometric KYC: Enhancing Ethical Compliance in U.S. Digital Banking [20], underscores the need for fairness-aware model design, integrating techniques such as sample re-weighting, threshold calibration, and cohort-based performance evaluation to ensure adaptive KYC systems remain both accurate and equitable.

Together, these advances mark a transition from static, document-based onboarding to dynamic, data-driven, and fairness-aware profiling. By combining adaptive scoring, entity linking, biometric verification, and fairness safeguards, AI has transformed KYC from a compliance formality into a continuously evolving capability that strengthens both regulatory alignment and institutional resilience against financial crime.

## 4     EFFECTS ON ACCURACY AND EFFICIENCY

The application of AI and machine learning in AML and fraud detection has produced measurable improvements in both accuracy of detection and efficiency of compliance operations. Traditional rule-based systems often generated false positive rates exceeding 95%, overwhelming investigators and diverting resources away from genuine threats. In contrast, modern AI models consistently demonstrate superior precision–recall performance, reducing unnecessary alerts while uncovering more complex typologies.

*Corresponding author

In transaction monitoring, hybrid approaches combining time-frequency features with machine learning classifiers have delivered significant gains. A Time-Frequency Based Suspicious Activity Detection for Anti-Money Laundering [7] achieved an AUC of 91.49% and reduced false positives to 11.85%, highlighting how statistical features such as kurtosis and skewness can strengthen discriminative power compared to handcrafted CRM rules. Similarly, ensemble models integrating XGBoost, LSTMs, and anomaly detection reached F1 scores above 0.90 with false positive rates below 3% [10], underscoring the leap in accuracy relative to scenario-based thresholds.

Graph-based learning methods have further advanced detection capability by capturing network motifs of laundering activity. Scalable Semi-Supervised Graph Learning Techniques for Anti-Money Laundering [21] outperformed pipeline ML by modeling temporal account interactions, while Regulatory Graphs and GenAI for Real-Time Transaction Monitoring [15] achieved F1 = 98.2%, precision = 97.8%, and recall = 97.0% on the Elliptic Bitcoin dataset. These results not only demonstrate improvements in accuracy but also introduce explainability through regulatory-grounded natural language justifications, thereby strengthening institutional trust in automated alerts.

Fraud detection has similarly benefited. Pattern recognition models for credit card transactions now outperform manual audits, while visualization techniques reveal hidden clusters that aid interpretability[13]. Reinforcement learning–based fraud defenses, though still experimental, show promise in dynamically adjusting detection strategies against adversarial behaviors, reducing both false negatives and operational costs [16].

Operational efficiency gains extend beyond detection to Suspicious Activity Reporting (SAR). NLP-driven SAR drafting systems reduce investigator workload by automatically extracting entities and transaction narratives [17]. When combined with retrieval-augmented generation (RAG), these systems accelerate report generation and ensure consistent alignment with regulatory standards, enabling faster case handling. In KYC, adaptive profiling and clustering reduce manual review burdens by dynamically prioritizing high-risk customers [19], while fairness-aware calibration prevents discriminatory biases that could otherwise result in regulatory pushback.

Taken together, these studies demonstrate that AI-driven AML systems achieve order-of-magnitude improvements over traditional rule-based approaches: false positive reductions of 50–90%, recall and F1 gains exceeding 20–30 percentage points, and faster investigative turnaround times across SAR and KYC. These outcomes illustrate that AI is not only improving the technical accuracy of compliance systems but also delivering efficiency benefits—reducing costs, shortening case closure times, and allowing compliance officers to focus on genuinely high-risk activity.

## 5 CONSTRAINTS AND CHALLENGES

Despite their clear advantages, AI applications in AML and fraud detection face significant constraints that limit widespread adoption. These challenges are not only technical but also organizational and regulatory, reflecting the complexity of embedding advanced models in highly regulated financial environments.

From a data perspective, the scarcity of high-quality labeled datasets is perhaps the most fundamental barrier. Confirmed cases of money laundering or fraud make up less than 1% of transaction data, creating severe class imbalance. Institutions are also constrained by strict confidentiality rules, which prevent sharing labeled data across organizations or jurisdictions. As a result, many studies rely on synthetic datasets or proprietary internal records [22]. This raises concerns about generalizability and real-world applicability. Privacy considerations further complicate matters: customer data is protected under regulations such as GDPR, limiting the ability to pool data for cross-institutional machine learning models[28].

At the model level, challenges center on explainability, robustness, and fairness. Black-box methods such as deep neural networks or graph neural networks can achieve high detection rates but often lack transparency, making it difficult for investigators and regulators to understand why a transaction was flagged [18]. Standardized, stress-tested evaluations of multimodal reasoning are therefore essential before deployment in regulated workflows, as evidenced by recent multidisciplinary benchmarks of MLLMs [26]. Bias is another concern: training data that underrepresents certain geographies or customer types can lead to unfair risk assessments, potentially exposing institutions to reputational and legal risks. Moreover, adversarial actors continuously adapt their tactics, raising the possibility that static ML models will be evaded or exploited over time.

Organizational barriers also slow adoption. Many financial institutions still rely on legacy systems built on relational databases and static rule engines. Integrating AI pipelines requires significant investment in data engineering, infrastructure, and staff training. Compliance analysts, accustomed to rule-based alerts, may initially distrust machine learning outputs, especially if explanations are limited. Cultural and skills gaps—between data scientists and compliance officers—can further hinder implementation.

*Corresponding author

Finally, regulatory challenges remain unresolved. Regulators demand models that are auditable, explainable, and consistent across jurisdictions. However, most AI techniques, particularly deep learning, struggle to provide explanations in regulator-accepted formats. Cross-border differences in AML regulations complicate deployment for global banks, while uncertainty over the acceptability of AI-driven SARs or automated customer profiling creates hesitation. Papers such as Enhancing Anti-Money Laundering Protocols: Employing Machine Learning to Minimise False Positives and Improve Operational Cost Efficiency [23] emphasize that regulatory acceptance remains a key bottleneck, despite strong technical results.

In sum, while AI has demonstrated substantial improvements in accuracy and efficiency, constraints in data access, model trustworthiness, organizational readiness, and regulatory acceptance continue to shape its trajectory. Addressing these challenges is critical if AI is to move from proof-of-concept projects to fully embedded compliance infrastructure.

## 6  FUTURE RESEARCH DIRECTIONS

While AI has already reshaped AML and fraud detection, its long-term potential depends on addressing current limitations and pursuing innovative research directions that align with compliance, operational, and regulatory realities. Several promising avenues have emerged.

First, the data bottleneck requires solutions that balance privacy with collaboration. Federated learning and privacy-preserving computation (e.g., secure multiparty computation, homomorphic encryption) allow institutions to train models collaboratively without sharing raw data. Additionally, machine unlearning that removes the influence of specific users or records from trained models—while balancing utility–privacy trade-offs—offers a compliance-friendly path for GDPR/CCPA deletion requests in AML/KYC settings (Li et al., 2025) [27]. This could unlock cross-institutional detection of laundering networks that span multiple banks while maintaining compliance with data protection regulations. Complementary to this, the creation of synthetic datasets and public AML benchmarks, carefully validated for realism, would reduce reliance on proprietary case studies and support reproducibility of research findings.

Second, advances in fairness-aware and explainable AI (XAI) are essential. Models must not only detect suspicious behavior but also justify decisions in ways that satisfy both investigators and regulators. Future research should focus on interpretable graph models, counterfactual explanations for SAR narratives, and cohort-based fairness testing to ensure risk scoring does not inadvertently discriminate across demographic or geographic groups. Combining high-performing black-box models with explanation layers—for example, SHAP or attention visualization in graph neural networks—can bridge the gap between accuracy and trust.

Third, there is scope for adaptive and adversarial robust detection. Criminals continuously evolve tactics, from layering through cryptocurrencies to exploiting trade-based laundering schemes. Reinforcement learning and multi-agent systems hold potential for proactive detection, allowing AML systems to simulate adversarial behavior and adapt monitoring thresholds in near real time. Similarly, time-aware models that incorporate sequential patterns (e.g., recurrent GNNs, temporal transformers) could capture laundering cycles and "slow smurfing" techniques that evade static detection.

Fourth, greater integration of generative AI and automation could streamline compliance workflows. Retrieval-augmented generation (RAG) systems show promise in automating SAR drafting with regulatory grounding, but future research must ensure consistency, multilingual support, and auditability before deployment in live Financial Intelligence Units (FIUs)[29]. Expanding these methods to KYC—such as dynamically updating customer risk narratives—could further reduce analyst workloads. Practical deployment will also benefit from synergized data-efficiency and compression techniques that retain utility while shrinking the LLM footprint [24].

Finally, visualization and human-in-the-loop systems remain underexplored. Interactive dashboards that combine graph analytics, anomaly scoring, and investigator feedback can help bridge machine detection with human expertise[30]. Huang et al. (2025) propose an immersive AR pipeline that fuses spatial scene understanding with hand-gesture recognition. Their real-time mapping + gesture control paradigm can translate to AML analyst tooling—for example AR overlays for graph motifs, alert triage, and evidence pinning. This line of work points to lower cognitive load and faster human-in-the-loop review in regulated workflows. Future research should study how analysts interact with these tools, ensuring usability and minimizing cognitive overload.

Overall, the field is moving toward next-generation AML systems that are collaborative, explainable, adaptive, and regulator-aligned. By combining federated learning, fairness-aware modeling, adversarial robustness, generative AI, and human-in-the-loop design, future research can transform AML from a compliance obligation into a real-time intelligence capability.

## 7  AI Application: Graph RAG + KYC

Building on the findings discussed above, this study extends the application of retrieval-augmented generative AI (Graph RAG) to enhance Know Your Customer (KYC) customer due diligence (CDD) in the final part of the paper. Traditional customer due diligence remains hindered by fragmented data sources, manual lookups, and inconsistent analytical workflows. Retrieval-augmented generation (RAG), a technique increasingly adopted for information retrieval and contextual reasoning, is well-suited to address these challenges in the KYC domain.

We propose to construct a RAG-based knowledge graph for bank customers by integrating both structured data (e.g., relational databases from core banking systems) and unstructured data (e.g., customer documents and reports). By linking the unified knowledge graph with a Large Language Model (LLM), which can interpret analyst queries and automatically translate to cypher language to retrieve relevant entities and relationships from the knowledge graph, compliance officers are able to access customer profiles and transaction relationships through natural language queries—without manual coding or complex navigation. This integration automates the generation and summarization of due diligence reports, reducing manual workload while improving consistency, traceability, and investigative efficiency. In addition, graph-based modeling provides rich contextual information, enhancing entity resolution and relationship visualization by allowing investigators to intuitively trace complex connections across customers, accounts, and counterparties. The graph data model also enables dynamic data updates, ensuring that new transactions, relationships, and risk indicators are continuously reflected in the knowledge graph. This capability provides a real-time, holistic view of customer risk, improves operational efficiency, and empowers compliance teams to focus on higher-value analytical tasks—such as comprehensive risk assessment and the early identification of potential financial crime patterns.

Importantly, while this proposal demonstrates the technical feasibility of a GraphRAG-based KYC assistant, real-world deployment would require additional governance measures—including persistent audit trails, human review, and model risk controls—to satisfy regulatory expectations under FATF and GDPR. In addition, although federated learning and privacy-preserving collaboration are noted as future directions, the current RAG-Graph proposal assumes centralized access to a synthetic financial knowledge graph, and does not aim to model cross-institution data sovereignty or jurisdictional privacy restrictions. Exploring these extensions is beyond the scope of the proposal but is a promising direction, which require additional mechanisms such as federated graph querying, secure multi-party computation, or differential privacy to enable collaborative risk analysis without exposing raw customer data.

### 7.1  RAG Agent Experiment Setting

The RAG Agent is structured around five primary components: Graph RAG Core, MCP Server, LLM Reasoning Loop, Configuration Layer, and Interfaces & Workflows (Fig. 2).

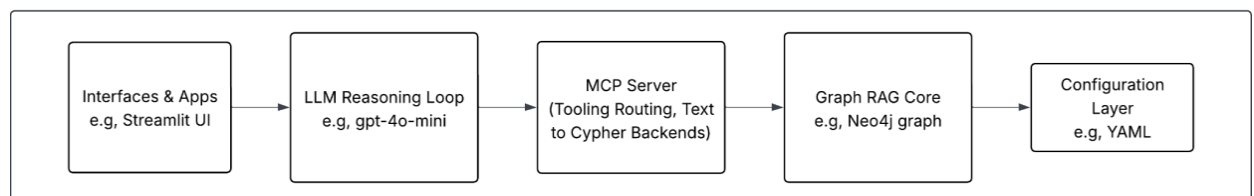

Figure 2 Agent Primary Components

#### 7.1.1  RAG Graph Core

The RAG graph (neo4j) serves as the primary knowledge base for the agent. All customer facts, accounts, transactions, sanctions, PEP links, alerts, and relationships stored in the RAG graph. Figure 3 shows the RAG graph database schema design. In this graph, each object in the KYC customer due diligence process (e.g., Customer,

*Corresponding author

Account, Transaction, Address, Sanction, PEP, Alert, CDD Case) is modeled as a node, while edges encode meaningful compliance relationships such as ownership, money flow, identity linkage, sanctions/PEP matches, shared identifiers, and investigative steps. This structure allows the RAG agent to traverse both behavioral patterns (multi-hop transaction paths) and contextual relationships (co-address, co-phone, shared counterparties), which are typically essential in Enhanced Due Diligence.

Graph construction is performed through idempotent MERGE-based ingestion, where new events (transactions, alerts, screening results) are incorporated as they arrive, and customer profile attributes or documents are refreshed through scheduled batch updates. This enables the graph to evolve dynamically without node duplication. For retrieval, only updated nodes trigger selective embedding refreshes, ensuring alignment between the RAG index and the live graph while avoiding full re-indexing.

By combining typed nodes, typed edges, and continuous updates, the graph database provides a structured and up-to-date representation of financial behavior that the RAG-Graph agent can query for multi-hop reasoning, pattern detection, and narrative KYC/EDD explanations.

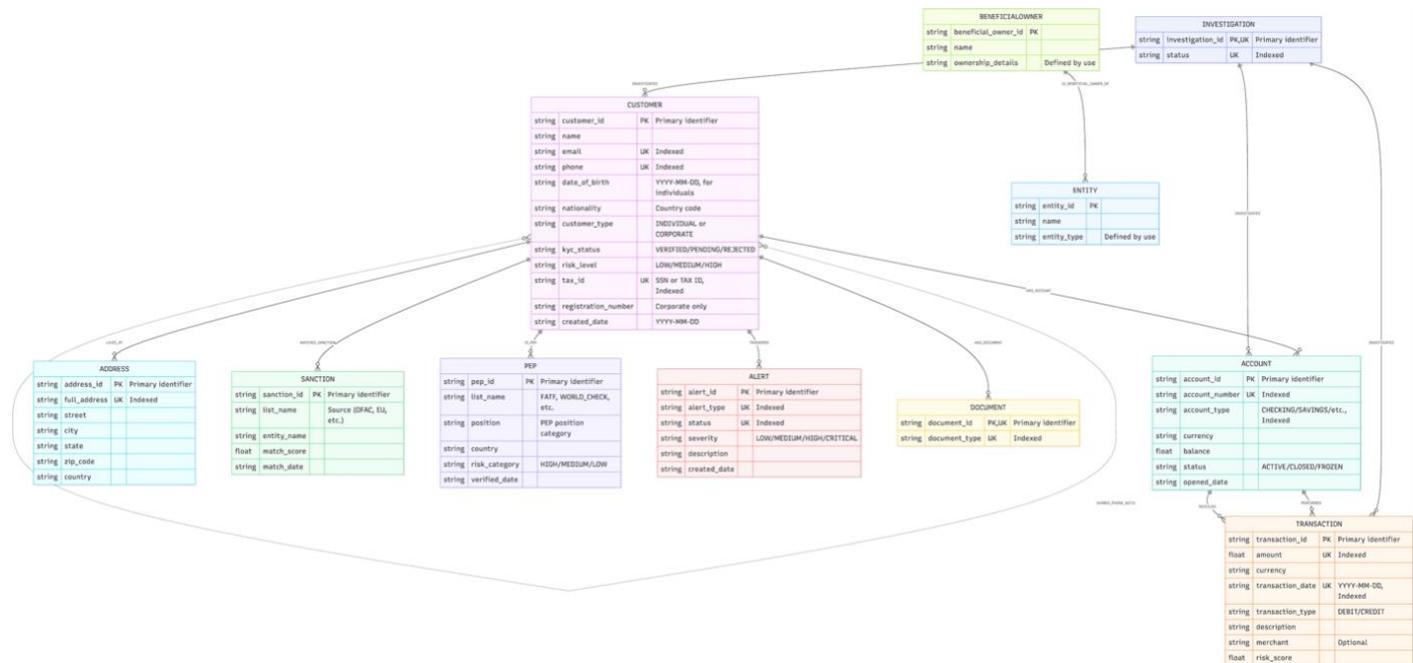

Figure 3. RAG Graph Database Schema Design

### 7.1.2 MCP Server

MCP is the middle layer between the LLM loop and the KYC graph: it advertises what the agent can do, executes those operations reliably, and feeds the polished results back into the reasoning loop.

The MCP toolset provides a suite of twelve callable functions that support end-to-end KYC due diligence and risk analysis, including: 1) execute_cypher: executes an explicit Cypher query with optional parameters and returns the raw Neo4j results. This function is used when the exact query is already known and does not require translation; 2) text_to_cypher: converts a natural-language question into a Cypher query using the graph schema, executes it, and returns both the generated query and its results; 3) get_customer_profile: retrieves a comprehensive customer dossier, including basic information, addresses, accounts, related parties, sanctions or PEP hits, alerts, and high-risk transaction counts; 4) get_customer_risk_summary: returns a focused risk snapshot that includes overall risk level, high-risk factors, sanctions and PEP details, and counts of transactions involving high-risk jurisdictions; 5) get_customer_accounts: provides customer identification details, name, total number of accounts, and per-account information such as account number, type, balance, status, and associated products; 6) get_customer_sanctions: retrieves sanctions, PEP, or watchlist exposure details for a single customer, including matched entities and supporting evidence; 7) find_customer_rings: detects potential suspicious networks by identifying customer rings connected through shared addresses, phone numbers, transactions, or relationships, and returns metadata and member lists for each ring; 8) extract_customer_transactions: extracts a customer's transaction records over a specified look-back window, including transaction risk scores, merchant details, and account context, along with summary statistics; 9) trace_shared_accounts: identifies other customers who share accounts with the target individual and flags any related sanctions or PEP exposure; 10) find_mutual_counterparties: discovers counterparties that have transacted with multiple specified customers within a given time window, useful for detecting shared transactional links or collusion patterns; 11) aggregate_customer_transactions: aggregates a customer's transactions by time interval (daily, weekly, or monthly) over a defined look-back period, returning both total amounts and transaction counts; 12) summarize_customer_risk_profile: generates a comprehensive, all-in-one risk summary combining account activity, relational links, transaction behavior, and compliance signals such as sanctions, PEPs, and alerts.

### 7.1.3 LLM Reasoning

The LLM layer (GPT-4o-mini) is "the brain" of the agent: it runs the conversation, chooses when to talk to MCP tools, and stitches the evidence into compliance-grade answers.

In this layer, we construct a system prompt (Figure 4) to guardrail the agent's persona, priorities, and answer format, include:1) who it is: an expert KYC investigator helping compliance analysts, so replies stay professional, risk aware, and grounded in the KYC domain; 2) How to behave: always answer the question directly first, cite data from tools, avoid speculation, highlight missing info—this keeps responses audit-friendly; 3) how to use tools: it reminds the model that the Neo4j MCP tools are the only source of truth and that it must call them when specific data is needed; 4) How to structure output: it enforces the "Direct Answer → Supporting Details → Key Findings" template, yielding consistent, scannable reports for downstream reviewers.

### 7.1.4 User Interface

The Streamlit interface provides compliance analysts with full control over the agent's execution environment, enabling them to select the LLM provider, launch natural-language investigations, and monitor which tools are invoked. Analysts can review structured outputs and run targeted workflows such as customer lookups, transaction tracing, and graph-based visualizations through an intuitive dashboard.

### 7.2 RAG Agent Performance Evaluation
### 7.2.1 Synthetic Evaluation Dataset

Due to privacy and regulatory constraints, KYC knowledge graphs containing personally identifiable information (PII) are not publicly accessible, and no public benchmark datasets exist for evaluating graph-based KYC customer due diligence systems. To ensure data availability while preserving confidentiality, we construct a synthetic dataset that reflects realistic customer attributes, relationships, and transactional patterns without exposing any sensitive information. The synthetic dataset used in this study consists of approximately 10,000 customer nodes, each with complete KYC profiles including risk levels, high-risk jurisdiction counts, and compliance flags. These customers collectively hold about 20,000–35,000 accounts (one to four per customer), generating roughly 250,000–500,000 transactions—each tagged with PERFORMED and RECEIVED relationships and enriched with counterparty information. The dataset also includes around 15,000–18,000 unique address nodes, with customers sharing addresses where appropriate. In terms of compliance artifacts, approximately 2% of customers match a sanction list, 1% are linked to politically exposed persons (PEPs), and 5% trigger one or more alerts. Finally, the graph contains diverse network relationships—including RELATED_TO, SHARES_ADDRESS_WITH, SHARES_PHONE_WITH, and counterparty-derived links—enabling both multi-hop reasoning and temporal behavior analysis within the KYC investigation framework.

*Corresponding author

```
SYSTEM_PROMPT = """You are an expert KYC (Know Your Customer) investigation agent assisting
compliance analysts in a bank.

Your role is to help analysts investigate customers by:

1. Answering questions about customer profiles, risk indicators, and relationships
2. Identifying potential compliance issues (sanctions, PEP status, suspicious activities)
3. Analyzing transaction patterns and network relationships
4. Providing clear, actionable insights for further investigation

You have access to tools through the Neo4j MCP Cypher Server. Use these tools to query the
graph database when you need specific information.

CRITICAL INSTRUCTIONS FOR ANSWERING QUESTIONS:
- **ALWAYS directly answer the question first** - State the answer clearly and concisely at
the beginning
- **Use only information from tool results** - Do not make assumptions or add information
not in the retrieved data
- **Cite specific data points** - Reference exact values from the tool results (e.g.,
"Customer has risk level MEDIUM" not "Customer has some risk")
- **Structure your response** - Start with a direct answer, then provide supporting details
- **Be concise and relevant** - Focus only on information that directly addresses the
question
- **If data is missing** - Explicitly state what information is not available rather than
inferring

Response Format:
1. Direct Answer: [One clear sentence directly answering the question]
2. Supporting Details: [Specific data points from tool results]
3. Key Findings: [Relevant insights only]

Example:
Question: "What is the risk profile of CUST000001?"
Good Answer: "Customer CUST000001 has a MEDIUM risk level. The profile shows:
risk_level='MEDIUM', 2 sanctions matches, 1 PEP flag, and 5 alerts. High-risk factors
include sanctions list matches and PEP status."

Remember: Your responses help compliance analysts make critical decisions. Always be
thorough, accurate, and directly answer the question asked."""
```

Figure 4 RAG Agent System Prompt

### 7.2.2 Evaluation Question Generation

To systematically assess the reasoning capability and factual grounding of the proposed RAG Agent, we design a multi-level evaluation framework composed of question sets aligned with increasing levels of graph reasoning and generative synthesis. The evaluation follows a five-tier difficulty schema (Table 1), where each level corresponds to a specific reasoning challenge ranging from direct fact lookup to multi-hop behavioral interpretation. The distribution across levels is selected to approximate the mix of question types encountered in real-world KYC and Enhanced Due Diligence workflows.

A total of 200 evaluation questions were generated using GPT-4o-mini. To ensure realistic contextual grounding, the LLM is connected directly to the live Neo4j graph and samples customer-specific facts (e.g., profile attributes, transactional behaviors, sanctions/PEP indicators, and network relationships). These sampled facts serve as input for the LLM to create question–answer pairs that conform to the predefined difficulty definitions for Levels 1–5.

Ground-truth answers are derived directly from Neo4j. The LLM assists only by generating the Cypher query needed for each question, while the final answer is obtained by executing that query and formatting the resulting database output into a natural-language response. This ensures that ground-truth answers originate from authoritative graph data rather than model-generated text. All ground-truth answers are stored with associated metadata (difficulty level, question type, quality rating) and manually reviewed by a human evaluator to ensure factual accuracy and alignment with the intended reasoning complexity.

Each generated pair then undergoes automatic quality verification by a separate LLM-based judge [32], also using the OpenAI backend. The judge evaluates the pair using a structured rubric assessing: 1) format correctness, 2) factual consistency with graph data, and 3) alignment with the intended difficulty level. Only question–answer pairs satisfying all rubric criteria are retained. The resulting dataset provides a controlled, difficulty-balanced benchmark for evaluating reasoning, retrieval accuracy, and factual grounding in KYC/EDD tasks.

*Corresponding author

| Level | % of Total Questions | Purpose |
|---|---|---|
| Level 1 (Baseline factual accuracy) | 20% | Test the agent's ability to retrieve precise attributes and maintain RAG faithfulness when responding to direct factual prompts. Example Questions:<br><br>1) "What is the country of residence of Customer A?"<br>2) "List all accounts owned by Customer B." |
| Level 2 (Entity–relation understanding) | 25% | Evaluate contextual grounding and accurate identification of relationship types between entities.<br>Example Questions:<br>1) "Which counterparties has Customer A transacted with in the past 12 months?"<br>2) "Who shares an address with Customer B?" |
| Level 3 (Graph reasoning capability) | 25% | Test the agent's capacity for multi-step reasoning, context retention, and inference over extended graph structures. Example Questions:<br><br>1) "Identify customers indirectly connected to a sanctioned entity through shared accounts."<br>2) "Which customers form part of a 3-node transaction chain involving Company X?" |
| Level 4 (Temporal and behavioral analysis) | 15% | Evaluate the model's ability to interpret temporal patterns, aggregated transaction behavior, and trend shifts.<br>Example Questions:<br><br>1) "Which customers increased transaction frequency with offshore accounts in Q3 2025?"<br>2) "Find accounts that received more than five incoming transfers from high-risk jurisdictions in the past month." |
| Level 5 (Narrative synthesis and risk interpretation) | 15% | Assess the agent's generative reasoning—combining multiple evidence sources, contextualizing relationships, and articulating regulatory-aligned explanations. Example Questions:<br><br>1) "Explain potential links between Customer A and Customer C given shared addresses, mutual counterparties, and a common employer."<br>2) "Summarize the risk profile of Customer B based on transactions, related parties, and watchlist matches." |

Table 1 Five-Tier Difficulty Schema

### 7.2.3    Evaluation Metrics

Traditional metrics such as ROUGE or BLEU are designed for summarization and translation tasks and are not appropriate for multi-hop factual reasoning in KYC/EDD workflows. Therefore, this paper employs RAGAS [31], a framework specifically developed for evaluating retrieval-augmented generation systems. RAGAS focuses on factual grounding, retrieval quality, and answer correctness rather than surface-level lexical overlap. The framework provides several complementary metrics:

- Faithfulness (0.0-1.0): Measures whether the generated answer is grounded in the retrieved context and contains no hallucinations
- Answer Relevancy (0.0-1.0): Evaluates how relevant the answer is to the question asked
- Context Precision (0.0-1.0): Assesses the precision of retrieved context (how much of the retrieved context is relevant)

- Context Recall (0.0-1.0): Measures the recall of retrieved context (how much of the relevant context was retrieved)

### 7.2.3 Evaluation Baseline

Existing rule-based KYC customer due diligence relies on static thresholds, isolated alerts, and periodic risk-score refreshes. Analysts must manually compile information from disparate systems, increasing cognitive load and introducing the risk of missing important contextual signals. These systems lack the capability to reason across multi-hop relationships or dynamically integrate new behavioral data. Because rule-based systems do not generate natural-language answers or evidence-grounded retrievals, they serve as a qualitative baseline rather than a quantitative one.

For quantitative baseline, a traditional vector-based RAG is used. The vector-based RAG baseline flattens Neo4j graph data into standalone text documents, embeds them using a sentence-transformer model, and retrieves relevant documents through cosine similarity. The retrieval of this method is limited to top-K document matches and must rely solely on textual similarity. It captures semantic similarity but loses the underlying graph structure, including multi-hop relationships, shared identifiers, fund-flow paths, and indirect counterparties. Therefore vector RAG serves as a simpler retrieval baseline, while GraphRAG provides deeper reasoning grounded in explicit entity-relationship structures.

## 7.3 Result and Analysis
### 7.3.1 Performance of Graph RAG agent

Figure 5 summarizes the RAGAS evaluation metrics—Faithfulness, Answer Relevancy, Context Precision, and Context Recall—across the five reasoning levels defined in the question design framework. The results show a clear performance gradient as question complexity increases from factual retrieval (Level 1) to narrative synthesis (Level 5).

At Level 1, the agent achieves near-perfect performance across all metrics (Faithfulness = 0.95, Answer Relevancy = 0.98, Context Precision = 1.00, Context Recall = 1.00), confirming that the system reliably retrieves and grounds simple factual information from the graph database. Level 2 questions, which require one-hop relational reasoning, also yield strong results (average scores above 0.85), demonstrating effective handling of entity–relation understanding and schema-aligned query translation.

Performance begins to decline at Level 3, where multi-hop reasoning and contextual inference are required. Faithfulness (0.83) and Answer Relevancy (0.96) remain high, but Context Precision (0.70) and Context Recall (0.46) drop significantly—indicating partial evidence retrieval and some loss of context during synthesis. This trend continues into Levels 4 and 5, which involve temporal, behavioral, and narrative reasoning. While the agent maintains relatively high Faithfulness (0.84–0.87) and strong Answer Relevancy (0.73–0.93), Context Precision and Recall decrease to the 0.38–0.79 range, reflecting growing difficulty in reconstructing complex temporal and relational dependencies across multiple graph paths.

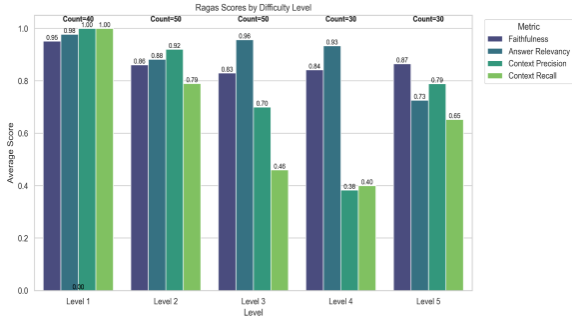

**Figure 5** RAGAS Result by Question Difficulty Level

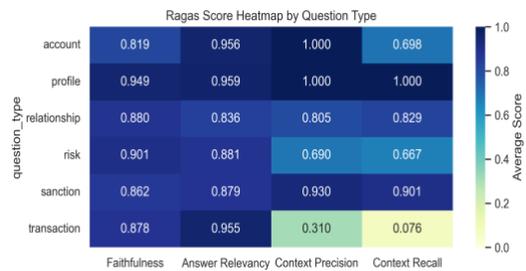

**Figure 6** RAGAS Result by Question Type

Figure 6 presents the RAGAS score heatmap, which analyzes model performance across different question types—account, profile, relationship, risk, sanction, and transaction. The agent performs best on profile and account queries, achieving perfect contextual grounding (Context Precision = 1.00; Context Recall = 1.00 and 0.70 respectively), reflecting strong retrieval for static attributes such as customer identity, nationality, and account ownership. Sanction-related questions also yield robust precision (0.93) and recall (0.90), indicating accurate linkage between customers and external compliance lists.

Performance moderately declines for relationship and risk queries, where reasoning depends on multi-hop connections or aggregated behavioral attributes. Here, Context Precision drops to 0.80 and 0.69, and Context Recall to 0.83 and 0.67, respectively—suggesting partial retrieval of relational and temporal evidence. The weakest results occur in transaction-type questions (Context Precision = 0.31; Context Recall = 0.08), where fine-grained temporal and counterparty details challenge the agent's retrieval depth and memory.

### 7.3.2 Graph RAG agent vs Existing Rule based System

The rule-based baseline reflects traditional KYC procedures in which analysts manually gather information from multiple disconnected systems and rely on static threshold rules to identify red flags. Because rule-based workflows do not perform automated retrieval or multi-hop reasoning—and do not generate natural-language answers—they serve as a qualitative baseline illustrating the limitations of static logic, fragmented data access, and analyst-dependent context gathering.

### 7.3.3 Graph RAG agent vs Traditional Vector RAG agent

Table 2 summarizes the experiment results between Graph RAG agent and traditional Vector RAG agent.

At Level 1, GraphRAG achieves near-perfect grounding (Faithfulness = 0.951; Answer Relevancy = 0.977; Context Precision = 1.00; Context Recall = 1.00), whereas Vector RAG performs poorly (Answer Relevancy = 0.042; Recall = 0.025). This indicates that even simple factual retrieval is difficult for embedding-only retrieval when the underlying data originates from a relational graph structure.

For Level 2, which requires modest relational reasoning, GraphRAG maintains strong contextual grounding (Context Precision = 0.92; Recall = 0.79). In contrast, Vector RAG retrieves only limited relevant evidence (Precision = 0.093; Recall = 0.14), highlighting the limitations of flattening graph data into static text documents.

Performance gaps widen significantly at Levels 3–5, where multi-hop traversal and behavioral reasoning are required. GraphRAG continues to exhibit high Faithfulness (0.83–0.87) and Answer Relevancy (0.726–0.957), while Vector RAG's relevancy collapses (0.03–0.123), and context completeness remains minimal (Recall = 0.06–0.333). These results demonstrate that Vector RAG cannot reliably reconstruct multi-entity relationships, as semantic similarity retrieval does not capture structural information such as counterparties, shared identifiers, fund flows, or temporal patterns.

| Question Level | Method | Faithfulness | Answer Relevancy | Context Precision | Context Recall |
|---|---|---|---|---|---|
| Level 1 | Graph_RAG | 0.951 | 0.977 | 1 | 1 |
| Level 1 | Vector_RAG | 0.525 | 0.042 | 0.105 | 0.025 |
| Level 2 | Graph_RAG | 0.861 | 0.882 | 0.92 | 0.79 |
| Level 2 | Vector_RAG | 0.637 | 0.123 | 0.093 | 0.14 |
| Level 3 | Graph_RAG | 0.83 | 0.957 | 0.7 | 0.46 |
| Level 3 | Vector_RAG | 0.475 | 0.03 | 0.067 | 0.08 |
| Level 4 | Graph_RAG | 0.842 | 0.934 | 0.383 | 0.4 |

|   | Vector_RAG | 0.466 | 0.096 | 0.033 | 0.333 |
|---|---|---|---|---|---|
|   | Graph_RAG | 0.865 | 0.726 | 0.789 | 0.653 |
| Level 5 | Vector_RAG | 0.779 | 0.061 | 0.065 | 0.094 |

**Table 2** Graph RAG Agent vs Traditional Vector RAG Agent Performance Comparison

Overall, these experimental and qualitative findings demonstrate that Graph RAG provides substantial improvements over both traditional rule-based workflows and embedding-based Vector RAG. Its ability to reason over graph structure, integrate diverse data sources, and produce grounded explanations positions Graph RAG as a significantly more effective foundation for automated KYC customer due diligence investigations.

## 8    CONCLUSION

The global challenges of money laundering and fraud require solutions that are accurate, scalable, and transparent. Traditional rule-based systems cannot match the speed and complexity of modern financial crime. As this review highlights, AI adoption across transaction monitoring, fraud detection, SAR, and KYC has significantly improved accuracy, reduced false positives, and enhanced investigative efficiency.

However, barriers such as data scarcity, explainability, and regulatory trust slow adoption. Financial institutions must strengthen data infrastructure and governance, while regulators should define clearer frameworks for AI auditability. Future progress lies in federated learning, adversarially robust models, and human-centered AI that balances automation with accountability.

Ultimately, the fusion of AML expertise with data-driven AI solutions and robust, scalable data infrastructure offers a path to build compliance systems that are efficient, interpretable, and resilient—transforming compliance from a reactive task into a proactive force for financial integrity. By reducing operational waste, enhancing transparency, and strengthening trust in digital financial ecosystems, AI-enabled AML systems also support broader goals of sustainable development in the digital economy.

*Corresponding author